\documentclass[sigconf]{acmart}
\usepackage{soul}
\usepackage{algorithm}
\usepackage{algpseudocode}
\usepackage{multirow}
\usepackage{caption}
\usepackage{subcaption}

\algdef{SE}[SUBALG]{Indent}{EndIndent}{}{\algorithmicend\ }%
\algtext*{Indent}
\algtext*{EndIndent}

\AtBeginDocument{%
  \providecommand\BibTeX{{%
    \normalfont B\kern-0.5em{\scshape i\kern-0.25em b}\kern-0.8em\TeX}}}

\begin{document}

\title[Improving Fairness in EHR Federated Learning]{Improving Fairness in AI Models on Electronic Health Records: The Case for Federated Learning Methods}

\author{Raphael Poulain}
\email{rpoulain@udel.edu}
\affiliation{%
  \institution{University of Delaware}
  \country{USA}
}
\author{Mirza Farhan Bin Tarek}
\email{mfarhan@udel.edu}
\affiliation{%
  \institution{University of Delaware}
  \country{USA}
}
\author{Rahmatollah Beheshti}
\email{rbi@udel.edu}
\affiliation{%
  \institution{University of Delaware}
  \country{USA}
}

\begin{abstract}
Developing AI tools that preserve fairness is of critical importance, specifically in high-stakes applications such as those in healthcare. However, health AI models' overall prediction performance is often prioritized over the possible biases such models could have. In this study, we show one possible approach to  mitigate bias concerns by having healthcare institutions collaborate through a federated learning paradigm (FL; which is a popular choice in healthcare settings). While FL methods with an emphasis on fairness have been previously proposed, their underlying model and local implementation techniques, as well as their possible applications to the healthcare domain remain widely underinvestigated. Therefore, we propose a comprehensive FL approach with adversarial debiasing and a fair aggregation method, suitable to various fairness metrics, in the healthcare domain where electronic health records are used. Not only our approach explicitly mitigates bias as part of the optimization process, but an FL-based paradigm would also implicitly help with addressing data imbalance and increasing the data size, offering a practical solution for healthcare applications. We empirically demonstrate our method's superior performance on multiple experiments simulating large-scale real-world scenarios and compare it to several baselines. Our method has achieved promising fairness performance with the lowest impact on overall discrimination performance (accuracy). Our code is available at \url{https://github.com/healthylaife/FairFedAvg}.
\end{abstract}

\ccsdesc[500]{Computing methodologies~Machine learning}
\ccsdesc[300]{Applied Computing~Life and medical sciences}

\keywords{Federated Learning, Algorithmic Fairness, Adversarial Fairness}

\maketitle

\section{Introduction}
The growing availability of electronic health records (EHRs, referring to the digitized collections of individuals' health information) offers great hopes in developing effective tools to inform prevention and treatment interventions across healthcare. 
The challenges of working with EHRs (such as large-scale, temporality, and missingness \citep{ehrchallenge}) present a ripe context for using machine learning (ML) methods dedicated to working with this special type of datasets. Similar to other types of human-generated and observational data, EHRs can also carry and be a source of various types of biases (especially historical and systematic ones) \citep{changfair}. In current healthcare ecosystems already dealing with longstanding concerns about health disparities (especially in places such as the US \citep{disp}), AI models that are created using EHRs can act as double-edged swords. While having huge potential to benefit some individuals; if not designed carefully, they can also unfairly impact others, especially underrepresented populations \citep{mimic_fairness}. Although studying justice (including fairness) in healthcare relates to a vast field, and a growing body of studies have looked at the fairness implications of AI models recently, dedicated studies to AI models in healthcare are limited \citep{fairhealth, equity}. The present study aims at presenting customized methods capable of addressing fairness in an important subset of AI health models.

As imbalanced training data is generally considered one of the primary roots of unfairness in AI models \citep{reviewfair}, having access to a larger dataset containing more samples from the smaller classes is an ideal way to achieve better learning opportunities for minority groups and thus improve fairness (broadly defined). In fact, existing studies targeting fairness mitigation in AI health models commonly assume having access to large datasets \citep{ehr_fair, pfohl_adv, correa}.  However, access to such data remains very limited in the case of EHRs, and sharing data across different sites faces numerous concerns about data privacy, security, and interoperability \citep{ehrprivacy}. Distributed learning paradigms, commonly referred to by federated learning (FL), are among the most popular choices in healthcare that can provide an adequate solution to alleviate such concerns. FL allows multiple healthcare sites to train a common large-scale model without sharing their local datasets. Investigating the fairness aspects of FL methods in EHR-based predictive models is the primary subset of AI health models that our study focuses on.

In a general FL setting, each site shares a common large-scale model without sharing its respective dataset. This design allows healthcare institutions to collaborate and jointly train a better model. While training on larger data accessible through multiple sites can help improve fairness (by increasing the number of samples in the minority groups), the non-IID data (IID: independent and identically distributed) across sites can lead to newer fairness issues. Some studies have addressed fairness in FL settings through local debiasing \citep{fade} and fairer aggregation strategies \citep{fairfed}, but the presented methods are often limited in several ways. First, fairness mitigation methods for FL settings often solely focus on the aggregation method on the server side \citep{fairfed}, leaving the local debiasing part to the discretion of each client (and unknown to the ML practitioners). Though this would allow for more flexibility on the client side, a unified debiasing method, where clients could build up on knowledge acquired by others, allows for better removal of sensitive information through a more diverse pool of training instances. Second, these methods assume a binary sensitive attribute (such as gender) \citep{fairfed, fade}. For example, the FairFed method \citep{fairfed} extended FedAvg\footnote{FedAvg is one of the most popular FL aggregation algorithms.} \citep{fedavg}, to alleviate group fairness concerns but solely considered binary sensitive attributes. However, studies have shown that non-binary attributes (race and insurance type), are usually sources of greater disparities in AI health models\citep{mimic_fairness}. It is therefore pivotal to study those biases and develop methods that are fair regardless of the source of biases. 

In this paper, we aim to address the aforementioned gaps in the literature by coupling local adversarial debiasing \citep{adv_fair} with an extension of FedAvg \citep{fedavg} and FairFed \citep{fairfed}. Specifically, we aim to implement techniques to improve fairness both at the local and aggregation level. In doing so, our study makes the following primary contributions:
\begin{itemize}
    \item We propose a comprehensive FL architecture with local adversarial debiasing of sensitive information in the context of healthcare where EHRs are used.
    \item We introduce a new fairness-metric-agnostic aggregation method for FL settings that can achieve better global fairness with non-binary sensitive attributes.
    \item{We demonstrate our method's performance by running an extensive series of experiments on multiple EHR datasets and comparing it to other baselines.}

\end{itemize}

\section{Problem formulation}
\label{section:background}
\label{section:fl}
\begin{figure}[th]
  \centering 
  \includegraphics[width=1\linewidth]{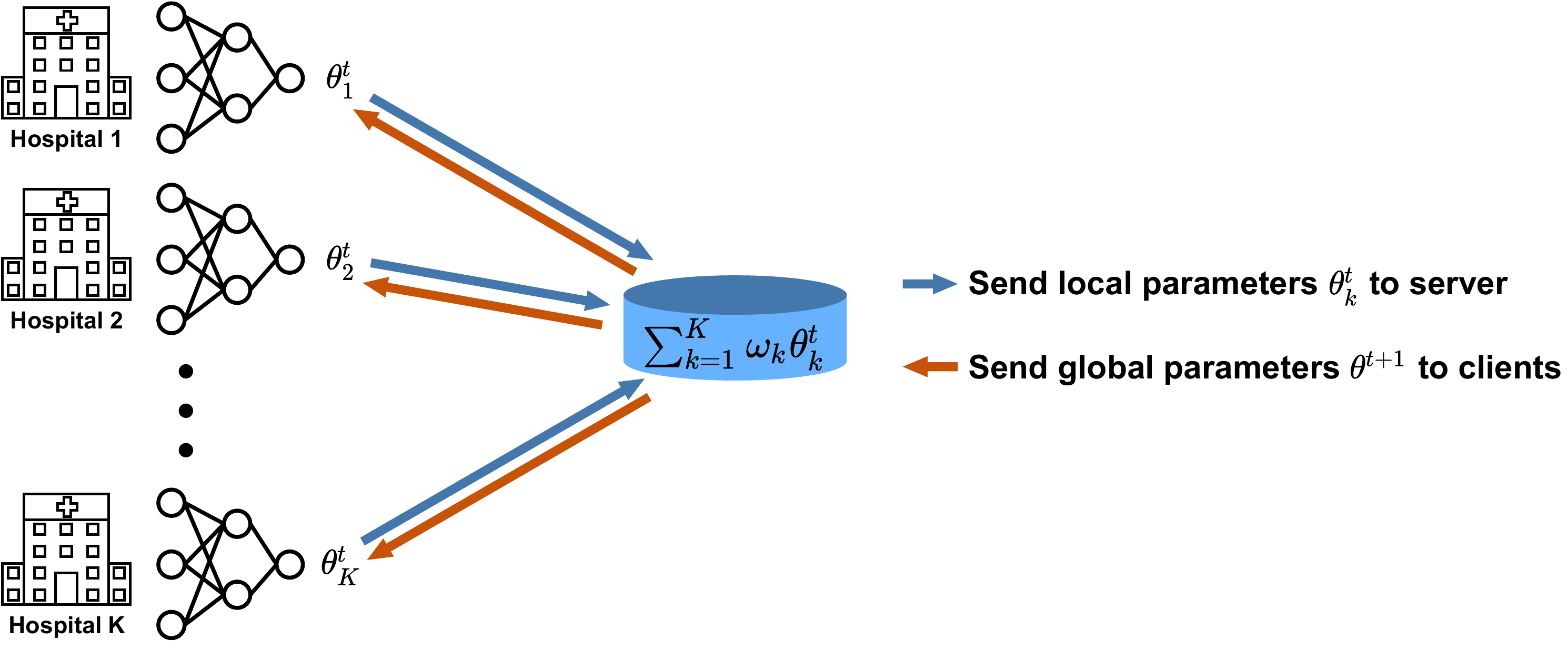} 

\caption{Federated Learning architecture where $K$ hospitals collaborate to learn a joint model. At each round, the clients send their trained parameters to the server (blue arrows). The server then aggregates them and sends the parameters back to the clients (orange arrows). The clients then update the model based on their local dataset.}
  \label{fig:fl_diag} 
\end{figure}

FL allows multiple clients (healthcare sites) to collaboratively train a single model with a set of parameters, shown with $\theta$ in our work. Amongst the many possible ways to formulate an FL problem, here, we describe it as a minimization problem of a weighted loss function across $K$ clients:
\begin{equation}
\min_{\theta}f(\theta) = \sum^K_{k=1} \omega_k  \mathcal{L}_k(\theta),
\label{eq:opt}
\end{equation}
where $\mathcal{L}_k(\theta)$ and $\omega_k$ are the loss and the weight of the $k$th client, respectively, such that  $\sum^K_{k=1} \omega_k = 1$.
As proposed by FedAvg \citep{fedavg}, a common workflow for an FL algorithm is based on the following steps: for each round $t$ of FL, the server randomly samples a set of clients and sends the lastly updated model's parameters $\theta^t$ to the clients. In parallel, each $k$ client will locally train the model starting from $\theta^t$ and send their newly updated parameters $\theta_k^{t+1}$ back to the server. Lastly, the server will aggregate the clients' parameters to form a new set of federated parameters $\theta^{t+1}$ such that $\theta^{t+1} =  \sum^K_{k=1} \omega_k \theta_k^{t}$. We provide a visual description of the FL process in a healthcare environment in Fig. \ref{fig:fl_diag}. We do not study peer-to-peer designs, as using client-server designs is more practical in real healthcare settings involving multiple sites, where usually a mutually trusted site can be considered as the server, reducing the complexities that peer-to-peer structures will add. 

To each $k \in K$ client, FedAvg assigns weights proportional to the size of their respective datasets, such that the client with the largest dataset has a greater impact on the overall optimization problem. Specifically, FedAvg defines each client's weight $\omega_k = \frac{n_k}{n}$, where $n_k$ is the number of instances in client $k$'s dataset, and $n$ is the total number of instances ($n = \sum^K_{k=1} n_k$). This weighting mechanism ensures that the global model is representative of the data distribution across all the participating clients. Though FedAvg provides great convergence guarantees \citep{fedavgconv}, its obliviousness to sensitive attributes might be a source of bias in the case where one of the clients with the largest assigned weight produces model parameters that are sources of strong biases \citep{fedfb}.

\textbf{Fairness Metrics:} Because of the sensitive nature of machine learning applications in healthcare and possible discriminating predictions made by such techniques, it is pivotal to define and quantify fairness metrics when evaluating new models. While our presented method is metric-agnostic supporting desired fairness metric, we focus on the metrics related to group fairness \citep{fairness_survey}. From the wide range of metrics used in the community, we focus on those that are most relevant to our study. Notably, we use parity-based metrics \citep{fairness_survey}, which measure the difference between the group outcomes, and minimax group fairness \citep {minimaxgroupfairness}, which measures the performance of the worst group outcome.

We use the  True Positive Rate (TPR) and Accuracy Parities, and the Worst-performing TPR. The TPR Parity requires that each group must have the same opportunity of being correctly classified as positive while the Accuracy Parity does not differentiate the positive and negative outcomes. In other words, each subgroup's TPR or accuracy should be independent of their sensitive attribute. Though it is clear how the absolute difference between the different groups' TPRs or accuracies can be used to quantify these parity-based metrics in the case of a binary sensitive attribute, it cannot be directly applied to non-binary sensitive attributes. A possible workaround would be to use the average difference across all possible subgroup pairs \citep{infofair}, however, using the standard deviation is easier to interpret. Therefore, we propose to use the True positive rate Parity Standard Deviation (TPSD), defined as the standard deviation of groups' TPR, and the Accuracy Parity Standard Deviation (APSD), defined as the standard deviation of subgroups' accuracy. Let us first note $A \in [\![1, N]\!]$ as a categorical sensitive attribute with $N$ possible values (race, for example), $Y \in [0, 1]$, and $\hat{Y} \in [0, 1]$ as the ground truth and the prediction of a binary predictor, respectively. We can now formulate the TPSD and APSD as follows:
\begin{equation}
    \label{eq:tpsd}
    TPSD = \sqrt{\frac{\sum^N_{n=1} (Pr(\hat{Y} = 1|A=n, Y=1) - \mu)^2}{N}}, 
\end{equation}

\begin{equation}
    APSD = \sqrt{\frac{\sum^N_{n=1} (Pr(\hat{Y} = Y|A=n) - \mu)^2}{N}},
    \label{eq:apsd}
\end{equation}
\noindent
where $\mu$ is the average TPR or accuracy across all groups. For the case of minimax group fairness, we note the worst performing TPR (the lowest) as Worst TPR and define it as follows:
\begin{equation}
    Worst TPR = \min_{n \in N} Pr(\hat{Y} = 1|A=n, Y=1).
    \label{eq:worstTPR}
\end{equation}
Note that, in the case of the parity-based metrics,  a smaller value is preferable, while a higher value is sought after for the Worst TPR.

\section{Method}
Our study builds upon three separate studies and integrates them into our comprehensive FL design. The first is the work described by Elazar et al. \citep{adv_fair}, where an adversarial approach was developed to remove demographic information from text representations. We couple this method with a second work offering an attention-based bidirectional RNN model, called Dipole \citep{dipole}. Dipole is one of the most studied EHR AI models and has been consistently used as a baseline for other studies. Using the method allows our FL design to capture the longitudinal intricacies of EHRs. The third method we adopt is FairFed \citep{fairfed} to aggregate clients' trained parameters. This method promotes better fairness guarantees than FedAvg, by allocating more weight to the clients with local fairness scores similar to the global ones. Inspired by this idea, we propose to allocate higher weights to the clients producing consistently fairer model parameters. Specifically, each client locally trains a Dipole-based model with adversarial debiasing on their local dataset and reports its fairness metric to the server. The server then aggregates the models' parameters by allocating more weight to the fairer clients.

\subsection{Dipole with Adversarial Debiasing}

\begin{figure*}[th]
  \centering 
  \includegraphics[width=0.35\linewidth]{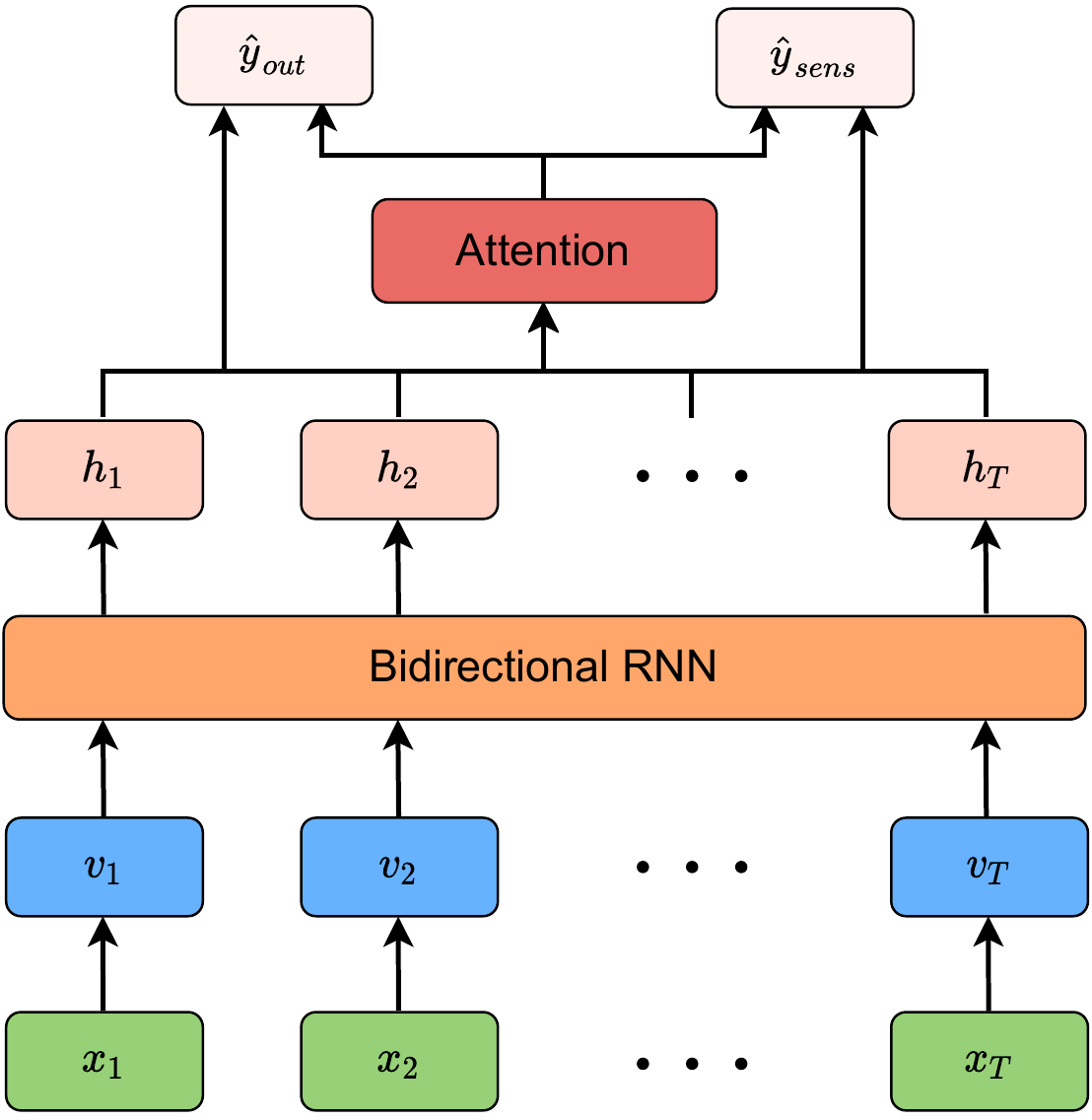} 

\caption{Dipole's architecture with adversarial debiasing.}
  \label{fig:architecture} 
\end{figure*} 

EHRs can contain a wide range of information about a patient's\footnote{The term \textit{patient} is broadly used and also includes healthy individuals interacting with healthcare systems for purposes such as prevention.} health, including demographic data, medical history, laboratory test results, and imaging reports. Without loss of generality, we present our method centered on a common subset that includes conditions, medications, and procedure codes in patients' health records. We note $C, M, P$ with sizes $|C|, |M|, |P|$, respectively as the set of the aforementioned codes. We also refer to $E = C + M + P$ as the set of all possible medical codes. Throughout this paper, we represent each patient $p \in P$ as a sequence of visits $x^p_1, x^p_2, ..., x^p_{n_v}$, where $n_v$ is the maximum number of visits and $x^p_t \in \{0, 1\}^{|E|}$ is a one-hot encoded vector representing the medical codes assigned to the $t^\text{th}$ visit for patient $p$.

We train a Dipole-based model with adversarial debiasing to learn sensitive information-free temporal representations. Accordingly, our model is composed of the following three main components: visit embeddings, bidirectional GRU, and an attention layer. We provide a visual representation of the entire architecture in Fig. \ref{fig:architecture}. For each visit, the one-hot encoded vector $x_t$ is fed to a Linear Layer with a rectified linear unit activation function (ReLU) to create the visit embeddings $v_t \in \mathbb{R}^m$ as $v_t = ReLU(W_v x_t + b)$, where $m$ is the embedding size, $W_v \in \mathbb{R}^{m * |E|}$ is a weight matrix, and $b \in \mathbb{R}^m$ the bias vector. The visit embeddings are then fed to a bidirectional GRU, where the input data is processed in a forward and backward manner: 
 $\overrightarrow{h_t},\overleftarrow{h_t}=GRU(v_t,h_{t-1})$, where $h_{t-1}$ is the previous hidden state, and $\overrightarrow{h_t} \in \mathbb{R}^h$, and $\overleftarrow{h_t} \in \mathbb{R}^h$ are the hidden states produces by the GRU layer for visit $t$ in the forward and backward direction, respectively, and $h$ is the size of the GRU's hidden states. We define the $t^\text{th}$ temporal visit representation as $h_t \in \mathbb{R}^{2h}$ derived from $h_t=|\overrightarrow{h_t},\overleftarrow{h_t}|$. We also note $H$ to be the vector containing all the temporal representations. 
 
Additionally, we adopt the general attention mechanism described in Dipole's original paper \citep{dipole}. 
This attention layer is trained to create a context vector $c_t \in \mathbb{R}^{2p}$ for the $t^\text{th}$ visit from the temporal visit representations $H$: $c_t = Attention(h_t, H)$.
To depict an accurate representation of patient $p$ at the $t^\text{th}$ visit, the hidden state $h_t$ and the context vector $c_t$  are concatenated and fed into a Linear Layer with a $tanh$ activation function: $r_t = tanh(W_r |h_t, c_t|)$.

We use this representation to perform two tasks: a (binary) outcome prediction task, and an adversarial debiasing following the ideas proposed in \citep{adv_fair}. Specifically, we train two classifiers where one would perform the binary prediction task and the second predict the patients' respective sensitive attributes from $r_t$. We refer to the first classifier by ``$out$'' and the second by ``$sens$.'' The Dipole-based network introduced above is trained to produce representations that carry minimum sensitive information without important information loss for the downstream task. In other words, the network will be trained to minimize the first classifier's loss while maximizing the second's loss.

Let us define the first linear classifier, $out$, that will use the representation $r_t$ and be trained to perform the binary prediction task:
\[
\hat{y}_{out} = W_o r_t + b,
\]
 where $W_o \in \mathbb{R}^{1*r}$, with $r$ being the size of the representation $r_t$. Similarly, we train the second linear classifier $sens$ to predict the sensitive attribute of each patient:
\[
\hat{y}_{sens}  = W_s r_t + b,
\]
where $W_s \in \mathbb{R}^{s*r}$, with $s$ being the number of possible values for the sensitive attribute. Let us now define $out$'s loss $\mathcal{L}_{out}$ with the Binary Cross-Entropy Loss (BCE), and $sens$'s loss $\mathcal{L}_{sens}$ with the Cross-Entropy Loss (CE):
\[
\begin{gathered}
    \mathcal{L}_{out} = BCE( \hat{y}_{out}, y_{out})
    \\
    \mathcal{L}_{sens} = CE( \hat{y}_{sens}, y_{sens})
\end{gathered}
\]
To ensure that $r_t$ is free of sensitive information, we train our network following the adversarial approach presented by Elazar et al. \citep{adv_fair}.
That is, we simultaneously train $sens$ to solely classify each patient based on their respective sensitive attribute while the rest of the Dipole-based network is trained to produce representations that $sens$ cannot classify and perform an outcome prediction task. Therefore, we define the network's loss $\mathcal{L}_{Dip}$ as:
\[
\mathcal{L}_{Dip} = (1-\alpha) \mathcal{L}_{out} - \alpha \mathcal{L}_{sens},
\]
where $\alpha$ is a hyperparameter that lets each client adjust for the impact of the local debiasing, allowing them to opt out of the adversarial part ($\alpha = 0$) of the training at any time, similar to what was proposed in FADE \citep{fade}.
This loss function will force the network to try to make it hard for $sens$ to classify each patient into their respective sensitive attribute (maximize $\mathcal{L}_{sens}$) while performing a prediction task.

\subsection{Fair Aggregation with non-binary sensitive attribute}
To ensure our model remains fair and accurate in an FL environment, we build upon the ideas proposed by FedAvg \citep{fedavg} and FairFed \citep{fairfed} and propose a new aggregation method of locally trained parameters that promotes global fairness. FedAvg solves the optimization problem described in Eq. \ref{eq:opt} by assigning a larger weight to the clients with a larger dataset. While this approach has been effective, it can lead to potentially severe biases if the clients with the largest datasets provide a biased set of parameters. Therefore, we propose to incorporate a fairness component in the aggregation process. Specifically, we assign the clients' weights based on a linear combination of their previous weight and their fairness metric, with the first round of weights assigned following the FedAvg method. Therefore, we assign greater weight to the consistently fairer and larger clients to push towards greater global fairness. To keep the proposed fairness mitigation method ``metric-agnostic'', we use a generic term $F$ to refer to the fairness metric to be optimized. Therefore, after initially assigning each client $k$'s weight $\omega_k^0$ with the original FedAvg formula,
\begin{equation}
\label{eq:ini}
\omega_k^0 = \frac{n_k}{\sum^K_{k=1} n_k},
\end{equation}
\noindent
for each round $t$, we update the clients' weights as follows:
\begin{align}
    \begin{split}
    \label{eq:weights}
        \Phi_k^t &= \begin{cases}
                    F_k^t, & \text{if $F_k^t$ is defined}\\
                    \frac{1}{K} \sum^K_{i=1}F_i^t, & \text{otherwise,}
                 \end{cases} 
        \\
        \omega_k^t &= \omega_k^{t-1} + \beta \left(\max_{i \in K}\Phi_i^t - \Phi_k^t \right),
    \end{split}
\end{align}

\noindent
where $F_i^t$ represents client $i$'s fairness score at round $t$, and $\beta$ is a hyperparameter to adjust the weight of the fairness component. When $F_k^t$ is undefined, we set $\Phi_k^t$ to be the average $F$ score across all clients, otherwise $\Phi_k^t = F_k^t$. Note that when $\beta = 0$, our method results in the original FedAvg, as the weights would remain unchanged throughout the rounds. Following this formula, the clients with the larger $F$ scores (poorer fairness) will receive a lower fairness reward as the difference between $\max_{i \in K}\Phi_i^t$ and  $\Phi_k^t$ will tend towards 0 and thus will have a lower impact on the parameters aggregation. This ensures that the global model becomes incrementally fairer as it will be  more impacted by the fairer set of parameters at any given aggregation round. Lastly, we ensure that $\sum^K_{k=1} \omega_k^t = 1$:
\begin{equation}
\label{eq:weight_div}
\omega_k^t = \frac{\omega_k^t}{\sum^K_{i=1} \omega_i^t}.
\end{equation}
We can now define the new set of the model's parameters as the weighted sum of the locally trained parameters:
\begin{equation}
\label{eq:upd}
\theta^{t+1} =  \sum^K_{k=1} \omega_k^t \theta_k^{t}.
\end{equation}
Following this method, each client solely has to communicate its $F$ score alongside its respective number of training instances which can be done without privacy concerns through methods such as Secure Aggregation \citep{secagg}. We provide a step-by-step description of the entire process in Algorithm \ref{alg:fairalg}.

\begin{algorithm}
\caption{Proposed Algorithm}\label{alg:fairalg}
\begin{algorithmic}
\State{\textbf{Server executes:}}
\Indent
    \State{Initialize model's parameters $\theta^0$}
    \State{Initialize clients' weights $\omega_k^0 = \frac{n_k}{\sum^K_{k=1} n_k}$}
    \For{each round $t=1, 2, ..., T$}
        \For{each client $k \in K$ \textbf{in parallel}}
            \State{$\theta^t_k, F^t_k \gets$ ClientUpdate(k, $\theta^{t-1}$)}
        \EndFor
        \State{$\omega_k^t = \omega_k^{t-1} + \beta \left(\max_{i \in K}\Phi_i^t - \Phi_k^t \right)$}
        \State{$\omega_k^t \gets \frac{\omega_k^t}{\sum^K_{i=1} \omega_i^t}$ }
        \State{$\theta^{t+1} \gets\sum^K_{k=1} \omega_k^t \theta_k^{t}$}
    \EndFor
\EndIndent
\newline
\State{\textbf{ClientUpdate($k, \theta$):}}
\Indent
    \State{$\theta_k^{t} \gets$ LocalTraining($\theta, X_k, Y_k$)} \Comment{$X_k, Y_k$ refer to client $k$'s local dataset.}
    \State{$F_k^t \gets F(\hat{Y}_k, Y_k, A_k)$} \Comment{$\hat{Y}_k, Y_k, A_k$: prediction, ground truth, and sensitive attribute, respectively.}
    \State \Return {($\theta_k^{t}, F_k^t$) to the server}
\EndIndent
\end{algorithmic}
\end{algorithm}

\section{Experiments}
To demonstrate the efficacy of the proposed method, we have realized a series of experiments in a wide range of scenarios in the healthcare domain. We have implemented our model using the Flower framework \citep{flower}, a customizable FL framework supporting large-scale FL experiments on heterogeneous devices. 

Throughout our experiments, we evaluated our method on four cohorts from two different datasets. Specifically, we have used, 1) the Synthea dataset \citep{synthea}, a public synthetic EHR simulation program that we have used to generate two cohorts of patients from different US states, and 2) MIMIC-III \citep{mimic3}, the most popular real-world EHR dataset of patients admitted to the Intensive Care Unit (ICU) from the Beth Israel Deaconess Medical Center in Boston, USA to generate two cohorts, one IID (independent and identically distributed) and one non-IID. 

\textbf{Synthea data:} For each cohort from the Synthea dataset, we have generated a dataset where each state has a number of patients proportional to their respective real population (California has a larger dataset than Pennsylvania).  We investigated fairness in a scenario (Most Populous States) where the five most populous US states (California, Texas, Florida, New York, and Pennsylvania) would collaborate (as the FL clients) and another one (Heterogeneous States) where we chose five US states with fairly different demographic distributions (Maine, Mississippi, Hawaii, New Mexico, and Alaska) to simulate a more diverse pool of clients. We focus on race as our sensitive attribute in our experiments since insurance type can change multiple times over the course of an individual's lifetime. Our model aims at predicting the mortality prediction task following an in-patient (hospital) visit. We present more details about our cohort extraction in Appendix \ref{a-sec:synt}.

\textbf{MIMIC data:} We generated two different cohorts from the MIMIC-III dataset, where we  assign MIMIC-III patients to five different FL clients. For the non-IID cohort, we synthesize non-IDD clients through a Dirichlet distribution as proposed in \citep{fairfed, dirichlet}, while the IID cohort was randomly generated following a uniform distribution. Similar to the Synthea scenario, we focus on race as our sensitive attribute, and the outcome prediction task relates to the standard task studied in the literature \citep{benchmarkmimic} related to predicting the mortality of patients admitted to the ICU, during their stay. We present more details about these cohort extractions in Appendix \ref{a-sec:mimic} and provide more details and statistics about the four cohorts in Table \ref{table:dset}. 

\begin{table*}[htbp]
    \caption{Statistics for all four cohorts: Most Populous States, Heterogeneous States, MIMIC-III IID and MIMIC-III non-IID.}
    \resizebox{\textwidth}{!}{%
    \begin{tabular}{ccccccccccc}
        \toprule 
        Race/Ethnicity & \multicolumn{5}{c}{Most Populous States}              & \multicolumn{5}{c}{Heterogeneous States}\\
        \cmidrule(lr){2-6}
        \cmidrule(lr){7-11}
        & California & Texas & Florida & New York & Pennsylvania & Maine      & Mississippi      & Hawaii      & New Mexico     & Alaska      \\
        White & 2330       & 1988  & 2188   & 2126     & 2016         &   2807     &   3323     &   776     &  2227     &   893  \\
        Black & 223        & 368   & 446    & 451      & 258          &    39     &    2384    &    69    &   87    &    55\\
        Asian & 508        & 129   & 79    & 261      & 70          &    26     &    63    &    1039    &   48    &    76 \\
        Hispanic & 2114        & 1849   & 945    & 713     & 175           &    29    &    195    &    298    &   2453    &   88\\
        Other & 184        & 81   & 67   & 98      & 55           &     77   &     96   &    949    &   389    &   305  \\
        \midrule
         & \multicolumn{5}{c}{MIMIC-III IID}              & \multicolumn{5}{c}{MIMIC-III non-IID}\\
        \cmidrule(lr){2-6}
        \cmidrule(lr){7-11}
        & Client 1      & Client 2      & Client 3      & Client 4     & Client 5 & Client 1      & Client 2      & Client 3      & Client 4     & Client 5       \\
        White &   1456 & 2201 & 1230 & 3016 & 2384  & 2188 & 2024 & 2050 & 1669 & 2356  \\
        Black &    142 & 213 & 106 & 329 & 166  & 163 & 154 & 165 & 145 & 329\\
        Asian &  52 & 70 & 34 & 111 & 69 &  0 & 119 & 92 & 124 & 1 \\
        Hispanic &    74 & 86 & 43 & 120 & 99 &   0 & 0 & 148 & 274 & 0 \\
        Other & 429 & 631 & 334 & 833 & 629 & 650 & 791 & 523 & 462 & 430 \\
        \bottomrule
    \end{tabular}}
    \label{table:dset}
\end{table*}

\textbf{Baselines:} For all scenarios, we evaluate the performance of the proposed method by comparing it to the following baselines. Our experiments allow us to compare the proposed method to a wide range of baselines that are most relevant to our work in the context of group fairness.

\textit{No-Fed:} Clients are trained individually on their local datasets, without any collaboration through FL. 

\textit{FedAvg:} FL using FedAvg \cite{fedavg} to aggregate the model's parameters. FedAvg has been one of the most popular FL algorithms and has been used as a baseline extensively. This method does not address group fairness. 

\textit{FairFed:} FL with FairFed \citep{fairfed}. FairFed extends FedAvg to incorporate a fairness element in the aggregation algorithm. To provide a fair comparison, we optimize FairFed for the same fairness metrics as the proposed method to consider non-binary sensitive attributes.

We also experimented with the impact of different values for the weight of the fairness component $\beta$ in Eq. \ref{eq:weights}.

For each experiment, we report the average accuracy, TPSD, Worst TPR, and APSD (as defined in Eq. \ref{eq:tpsd} to \ref{eq:worstTPR}) and their standard deviation in a 5-fold cross-validation process. Note that for the No-Fed baseline, though the training and testing are done individually, we report the metrics collectively for all clients (as if the testing sets were centralized).

\section{Results}

\begin{table*}[htbp]
    \centering
    \caption{Results on all cohorts. Mean $\pm$ standard deviation. $\uparrow$: Higher is best, $\downarrow$: Lower is best.}
        \resizebox{\linewidth}{!}{%
        \begin{tabular}{cccccccc}
            \toprule
             &  & No-Fed &  FedAvg  & FairFed & Proposed \\
            \midrule
             \multirow{4}{*}{Most Populous States} 
             & Accuracy $\uparrow$           &     87.45 $\pm$ 0.46       &   \textbf{90.02 $\pm$ 0.41}    &   89.49 $\pm$ 0.41             &   89.28 $\pm$ 0.45            \\
             & TPSD $\downarrow$             &     0.184 $\pm$ 0.032      &   0.102 $\pm$ 0.037            &   0.066 $\pm$ 0.015            &   \textbf{0.052 $\pm$ 0.014}  \\
             & Worst TPR $\uparrow$          &     0.368 $\pm$ 0.047      &   0.441 $\pm$ 0.092            &   \textbf{0.509 $\pm$ 0.083}   &   0.502 $\pm$ 0.080           \\
             & APSD $\downarrow$             &     4.81 $\pm$ 1.19        &   2.88 $\pm$ 0.98              &   1.79 $\pm$ 0.66              &   \textbf{1.51 $\pm$ 0.62}    \\
            \midrule
             \multirow{4}{*}{Heterogeneous States}
             & Accuracy $\uparrow$           &     87.72 $\pm$ 0.26       &   \textbf{90.08 $\pm$ 0.30}    &   89.39 $\pm$ 0.32             &   89.68$\pm$ 0.34             \\
             & TPSD $\downarrow$             &     0.225 $\pm$ 0.023      &   0.064 $\pm$ 0.016            &   0.042 $\pm$ 0.021            &   \textbf{0.031 $\pm$ 0.022}   \\
             & Worst TPR $\uparrow$          &     0.258 $\pm$ 0.052      &   0.572 $\pm$ 0.066            &   0.062 $\pm$ 0.069            &   \textbf{0.639 $\pm$ 0.064}   \\
             & APSD $\downarrow$             &     6.26 $\pm$ 0.99        &   2.42 $\pm$ 0.37              &   1.37 $\pm$ 0.38              &   \textbf{1.16 $\pm$ 0.38}     \\
              \midrule
             \multirow{4}{*}{MIMIC-III IID}
             & Accuracy $\uparrow$           &     74.32 $\pm$ 0.96       &   \textbf{76.59 $\pm$ 0.55}    &   75.96 $\pm$ 0.64             &   76.38 $\pm$ 0.52             \\
             & TPSD $\downarrow$             &     0.088 $\pm$ 0.019      &   0.051 $\pm$ 0.014            &   0.035 $\pm$ 0.015            &   \textbf{0.034 $\pm$ 0.016}   \\
             & Worst TPR $\uparrow$          &     0.631 $\pm$ 0.057      &   0.689 $\pm$ 0.037            &   \textbf{0.749 $\pm$ 0.043}   &   0.734 $\pm$ 0.037            \\
             & APSD $\downarrow$             &     7.29 $\pm$ 0.20        &   3.57 $\pm$ 0.92              &   3.15 $\pm$ 0.82              &   \textbf{2.87 $\pm$ 0.85}     \\
             \midrule
             \multirow{4}{*}{MIMIC-III non-IID}
             & Accuracy $\uparrow$           &     74.38 $\pm$ 0.51       &   \textbf{77.05 $\pm$ 0.29}    &   76.49 $\pm$ 0.64             &   76.44 $\pm$ 0.54             \\
             & TPSD $\downarrow$             &     0.110 $\pm$ 0.014      &   0.049 $\pm$ 0.013            &   0.034 $\pm$ 0.014            &   \textbf{0.033 $\pm$ 0.014}   \\
             & Worst TPR $\uparrow$          &     0.644 $\pm$ 0.109      &   0.735 $\pm$ 0.081            &   0.763 $\pm$ 0.045   &   \textbf{0.784 $\pm$ 0.129}            \\
             & APSD $\downarrow$             &     7.01 $\pm$  1.45       &   3.35 $\pm$ 1.29              &   2.94 $\pm$ 0.68              &   \textbf{2.16 $\pm$ 1.06}     \\
            \bottomrule
        \end{tabular}}
        \label{table:results}
\end{table*}

Table \ref{table:results} shows the Accuracy, TPSD, Worst TPR, and APSD scores for all methods on all datasets when using adversarial debiasing. The proposed method is able to achieve competitive performance with an accuracy that remains close to the one achievable using FedAvg (i.e.,  without any fairness mitigation, other than the adversarial debiasing) while demonstrating better group fairness than all baselines tested for all but one metric (Worst TPR). Notably, our method achieves comparable accuracy to FairFed while reducing the parity-based metrics on all cohorts. Though FairFed seemed to outperform our proposed method for the minimax group fairness metric (Worst TPR), the results remain very close to each other. It is also worth noting that, as expected, using FL allows for both better predictive performance and fairer predictions, as shown by the poorer scores achieved with the local training. 

\begin{figure*}[th]
     \centering
     \begin{subfigure}{0.45\textwidth}
         \centering
         \includegraphics[width=\textwidth]{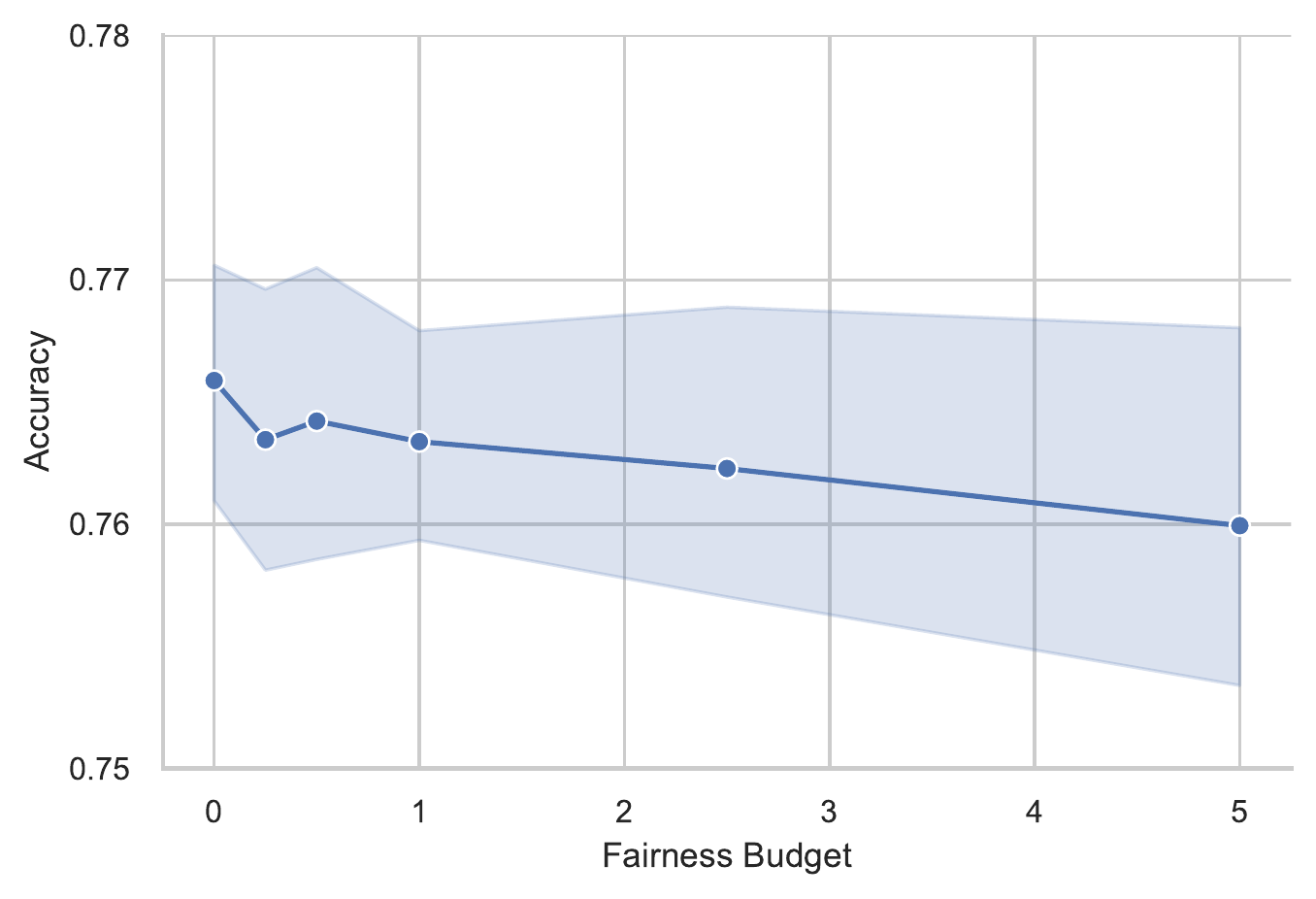}
     \end{subfigure}
     \begin{subfigure}{0.45\textwidth}
         \centering
         \includegraphics[width=\textwidth]{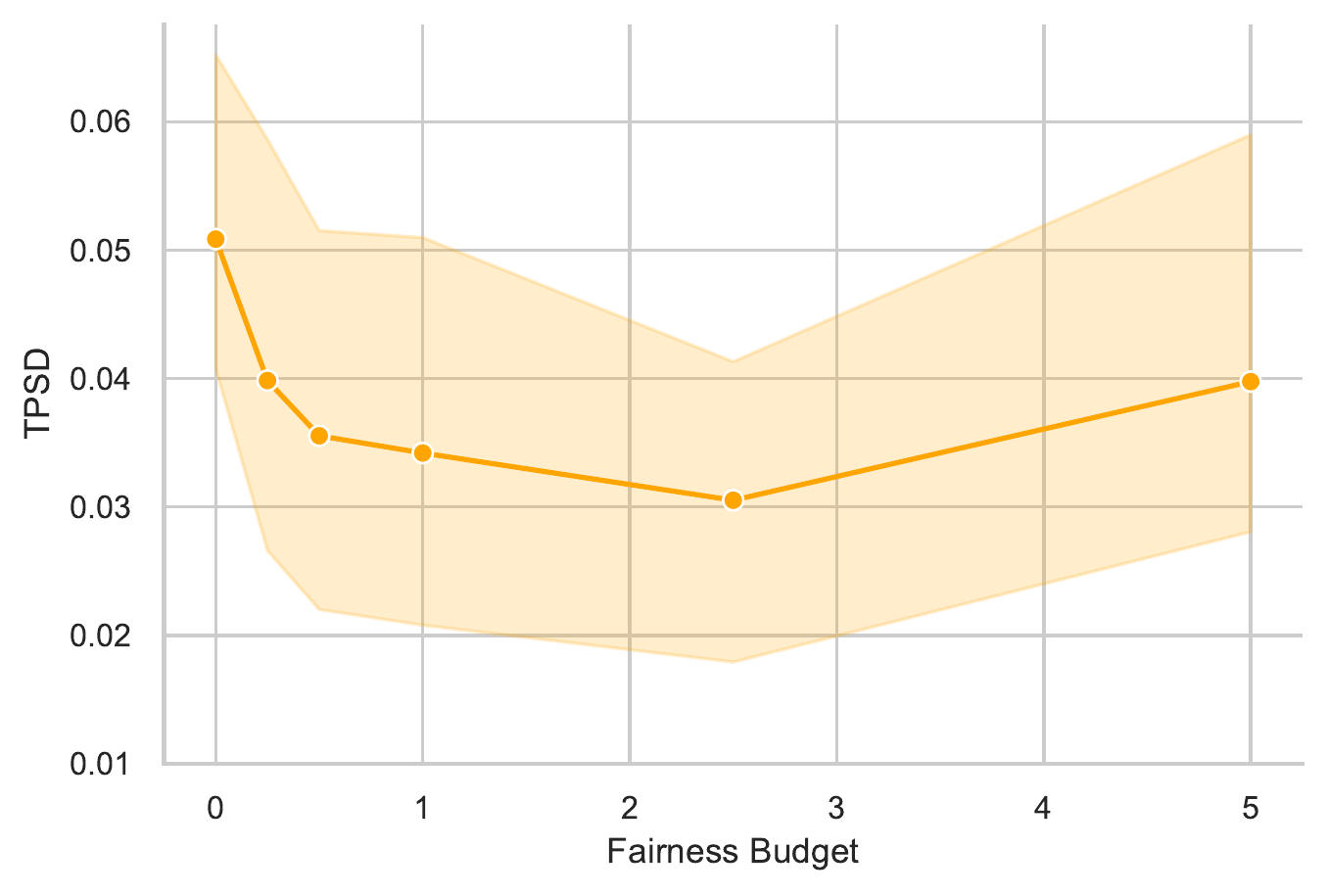}
     \end{subfigure}
     \begin{subfigure}{0.45\textwidth}
         \centering
         \includegraphics[width=\textwidth]{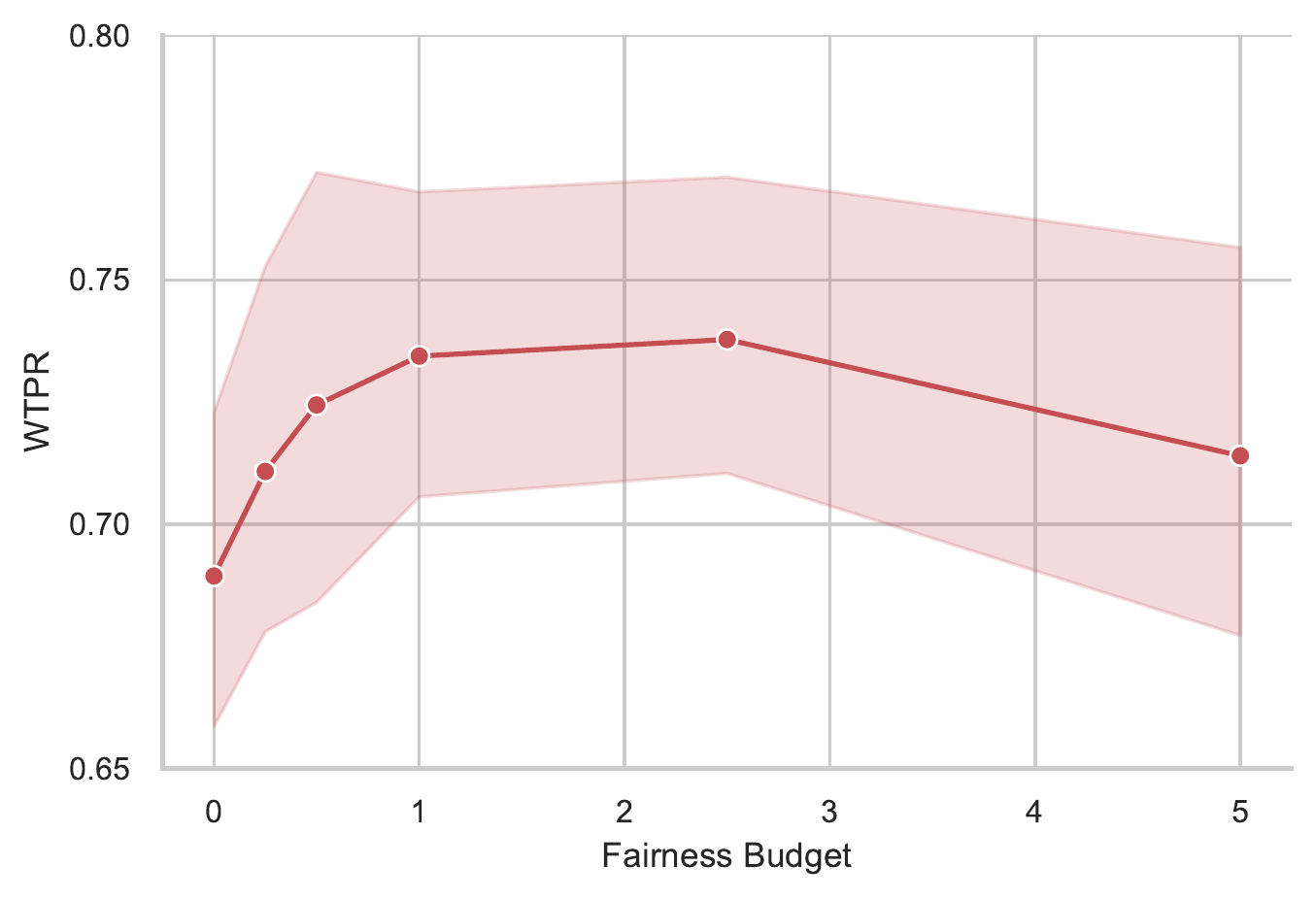}
     \end{subfigure}
     \begin{subfigure}{0.45\textwidth}
         \centering
         \includegraphics[width=\textwidth]{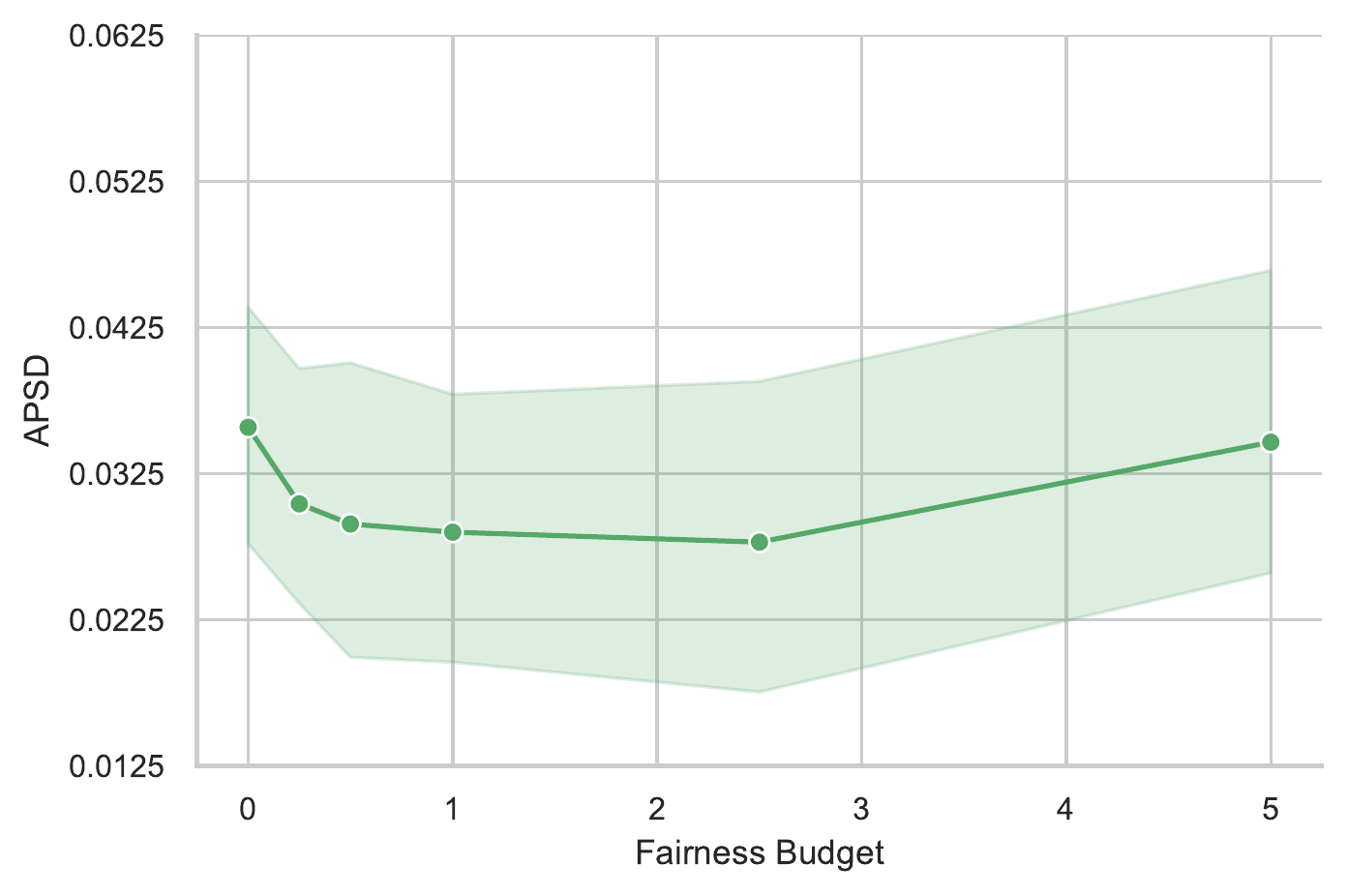}
     \end{subfigure}
        \caption{Impact of the fairness budget $\beta$ on the Accuracy, TPSD, Worst TPR, and APSD for the MIMIC-III IID cohort. The shaded areas represent the standard deviation.}
        \label{fig:fairness} 
\end{figure*}

\textbf{Impact of fairness  reward:} 
Similar to FairFed, we have implemented a fairness budget $\beta$ for the calculation of the clients' weights in the aggregation process. We study the impact of $\beta$ both on the fairness of the resulting predictions and the overall accuracy (Fig. \ref{fig:fairness}). We only report the results related to the MIMIC-III IID cohort, though we found similar results in other cohorts throughout our experiments. Specifically, we have set $\beta$ to different values (0, 0.25, 0.5, 1, 2.5, 5).  $\beta = 0$ corresponds to FedAvg, thus without a fair aggregation process.

As expected, increasing the fairness budget provides better fairness guarantees as shown by the differences in the fairness metrics when $\beta$ ranges from 0 to 2.5. One can notice a decrease in TPSD and APSD from 0.051 and 3.75 to 0.030 and 2.78, respectively, while the Worst TPR increased from 0.689 to 0.738. Note that for the parity-based measures (TPSD/APSD), a lower value is better while a higher value reflects better results for the accuracy and Worst TPR. Additionally, though it is hardly perceptible in the figure, the accuracy also decreases from 76.59 to 76.22. However, when $\beta = 5$, the method shows poor performance, both for the fairness metrics and the accuracy. This is most likely due to the fact that the fairness budget becomes too important in the aggregation process, making the clients' weights change drastically from one round to the other, and making the optimization process less stable.

\section{Related Work}

While there exists a wide range of studies related to our work, here, we discuss a non-exhaustive group of studies that are closely related to ours, including the applications of FL on EHRs, the studies investigating group fairness in predictive modeling with EHRs, and the works on predictive models under an FL setting focusing on improving group fairness.

\noindent
\textbf{Federated Learning with EHRs:} 
FL has been gaining significant attention in the healthcare sector, especially for its privacy-preserving nature and ability to work on decentralized data \citep{dang2022federated, fedhealthreview, nguyen2022federated}. FL has been applied to EHRs in various applications, with one of the earliest being in-hospital mortality prediction within patients admitted to the ICU (i.e., the prediction task used in our experiments). Multiple studies \citep{sharma2019preserving,pfohl} investigated the use of different machine learning models such as logistic regression, multi-layer perception (MLP), and feed-forward neural networks on the MIMIC-III \citep{mimic3} and eICU \citep{eicu} EHR datasets. These studies aimed to utilize FL to improve the accuracy of models while protecting patient privacy. To name a few of such studies, consider the autoencoder-based architecture proposed to cluster patients for mortality prediction \citep{autoencfl} using the eICU dataset. Another work investigated MLP and LASSO logistic regression in an FL setting to predict 7-day mortality for hospitalized COVID-19 patients with EHR data \citep{vaid2020federated}. While these studies have demonstrated the potential of FL in healthcare, the underlying AI models remain fairly simple and the possible fairness concerns are not investigated. To mitigate those concerns, recent studies proposed to use FL for improving fairness by increasing the dataset diversity \citep{algfarihealth} (without developing the idea further).

\noindent
\textbf{Improving Group Fairness on EHRs: }
Though group fairness remains to be investigated in FL with EHRs, bias mitigation techniques on EHRs have been studied for centralized models in a few studies \citep{ehr_fair}. Notably, Pfohl et al. \citep{pfohl_adv} proposed to use adversarial learning to debias a model to predict the risk of cardiovascular diseases from EHRs. A similar idea has been applied to disease prediction using chest X-ray \citep{correa}. Intra-processing techniques such as fine-tuning and pruning of pre-trained networks have also been proposed for chest X-rays \citep{pruning}. Additionally, regularization techniques have been used to achieve counterfactual fairness \citep{pfohl_constr}, where the model is required to output the same prediction for a patient and the same patient when changing the value of their sensitive attribute. Some studies also argued that, in critical applications such as healthcare where the trade-off between accuracy and fairness could lead to dramatic outcomes, fairness should only be achieved through data collection rather than model constraints, which can be indirectly achieved through FL \citep{chen}.

\noindent
\textbf{Group Fairness in FL:} 
While fairness in FL can refer to many different subfields, such as client-based fairness \citep{clientfair, wangfair, ditto, agnostic} (enforcing that the global model performs fairly amongst the clients), here, we focus on the studies investigating group fairness, which has seen growing interest. To name a few, Zeng et al. proposed FedFB \citep{fedfb} to extend FairBatch \citep{fairbatch} to an FL setting. FairFed \citep{fairfed} proposed a fairness-aware aggregation algorithm agnostic to the clients' local debiasing technique. Also, Abay et al. \citep{abay} investigated the effectiveness of a global reweighting strategy.
As opposed to these studies that mostly aim at reducing the difference between each group's outcomes, FedMinMax \citep{fedminmax} proposed an optimization algorithm to tackle minimax group fairness \citep{minimaxgroupfairness} in an FL setting. Similarly,  PFFL (Provably Fair Federated Learning) \cite{pffl} claims to improve group fairness in FL with a variation of Bounded Group Loss \citep{bgl} to ensure no group loss is below a given threshold. Similar to our work, Federated Adversarial DEbiasing (FADE) \citep{fade} proposed to leverage adversarial debiasing \citep{adversarialDebiasing} to enhance group fairness in a federated setting, however, FADE's aggregation process does not consider any fairness metric. While these studies have demonstrated promising results in group fairness, most of them solely consider fairness as a binary problem and are often dependent on a single fairness metric. Additionally, these methods either study the federated algorithm or the model training process, leaving the combination of both underinvestigated. In this study, we investigated a comprehensive FL approach, from the model and training process to the weights aggregation algorithm, to enhance fairness (agnostic of the chosen metric) in a healthcare setting.

\section{Discussion}
Though FL applications with EHRs have seen growing interest, fairness-related issues in this context remain widely under-investigated.  In this study, we investigated the fairness benefits of FL methods and proposed a new FL algorithm to leverage EHRs with better fairness guarantees for non-binary sensitive attributes (like race or insurance type). Our proposed algorithm aims at giving greater weight to consistently fairer clients to push toward better global fairness. Our main results in Table \ref{table:results} demonstrate that FL is a viable paradigm for healthcare sites to collaborate and train a common AI model to achieve better and fairer performance while alleviating privacy and security concerns. Additionally, the difference in fairness between the two cohorts with homogeneous data distributions (the Most Populous States and MIMIC-III), and the Heterogeneous States cohort indicates that an FL environment with heterogeneous clients provides even greater fairness guarantees. On the contrary, when using Local (No-Fed) training, the fairness performance gap is smaller for the homogeneous cohorts. Of particular note is the Heterogeneous States scenario, where the different demographics of each state allow the common model to have access to a much more diverse population which is key in training accurate and fair AI models.

We have added a hyperparameter $\beta$ to control the weight of the fairness component in the aggregation process to ensure that the clients' weights do not change too drastically between each round while keeping the model as fair as possible. Additionally, we have demonstrated the importance of the fairness budget in our aggregation process and how it can impact the model's accuracy and fairness guarantees. 


\textbf{Limitations:} Our study remains limited in a few ways. We have limited our study to a binary classifier. Also, our proposed method only considers one sensitive attribute at a time. We expect, however, that the same design would expand to non-binary scenarios or scenarios with multiple sensitive attributes as well. Moreover, we have only considered adversarial debiasing as our fairness mitigation technique, leaving other techniques to be studied in an FL context. Additionally, we have not studied the impact of the proposed method on a client level. It is possible that some clients' performance could be affected negatively to achieve better fairness. Lastly, though Dipole has been one of the most successful AI models in healthcare applications, newer transformer-based models may also fit certain applications better. We note, however, integrating large-scale transformer-based models into FL designs may be challenging in practice (due to the computational resources they would need). 

As for future works, one can investigate the impact of the proposed method in scenarios with a greater number of clients, and study additional fairness metrics, beyond group and worst-case fairness, as well as study further the impact of both hyperparameters $\alpha$ and $\beta$. Additionally, as we have not directly reported the performance of the proposed method for each individual subgroup, one can investigate the effect of the approach on each group individually in addition to the common metrics like TPSD.



\begin{acks}
Our study was partially supported by the NIH awards 3P20GM103446 and 5P20GM113125.
\end{acks}

\bibliographystyle{ACM-Reference-Format}
\bibliography{references}

\appendix
\section{Synthea dataset cohort extraction}
\label{a-sec:synt}
In this experiment, we simulated a generic scenario where the five most populous US states collaborate to jointly train a federated model and another with a selection of five heterogeneous states. Here, each state represents a client in the FL scope (or a hospital, following Fig. \ref{fig:fl_diag}). These selections allow us to investigate fairness in two realistic FL scenarios across the country, one centered on larger clients, and one to evaluate the proposed method's robustness to heterogeneous demographic distributions. 

We simulate a scenario where the states represent the clients of our FL architecture and thus would collaborate to learn a federated model. Therefore, the states' data is never shared with the central server nor with the other participating states. We have performed two separate experiments with their separate dataset generation processes to fit the needs of the desired scenario. Regarding our second cohort (Heterogeneous States), we have selected those states because of their distinctive data distribution. For example, Maine, Mississippi, Hawaii, New Mexico, and Alaska have the highest White, Black, Asian, Hispanic, and Native American populations per capita according to a 2020 US census \citep{census}, respectively. 

Note that the two cohort generation processes are independent of each other, the states' populations are only proportional within their own scenario. From these two cohorts, we aim to develop an accurate mortality prediction model for patients who have been discharged following an inpatient visit (excluding those who passed away during the inpatient visit) \citep{cehr_gan_bert} while preserving group fairness. The latest inpatient visit in the medical history of each patient is defined as the anchor event, leading to an observation window spanning the entire medical record up until the latest inpatient visit, and a prediction window of one year following the patient's latest inpatient visit. We extracted the condition, medication, and procedure codes during the observation window and used them as input to our model. As shown in \citep{mimic_fairness}, race and insurance type can be a source of bias for machine learning applications to healthcare and are thus most relevant to study in our work. 

\section{MIMIC Dataset Cohort extraction}
\label{a-sec:mimic}
We have created two separate cohorts from the MIMIC-III dataset where the patients are assigned to clients, representing a healthcare institution (states in the case of Synthea). From the original MIMIC dataset, we extracted two cohorts: a non-IID and an IID one. The non-IID cohort was generated following the ideas described in \citep{fairfed, dirichlet}, where we have simulated non-IID clients through a Dirichlet distribution with $\alpha = 1$ (when $\alpha\to\infty$, the data distribution would be IID). For the IID cohort, the number of patients per each of the five created clients is  randomly generated from a uniform distribution of 1,500 to 5,000 patients. From these cohorts, we aim to predict the mortality of patients admitted to the ICU, during their stay. The anchor event for this task is defined as an admission to the ICU. The observation window encompasses the first 24 hours of the patient's ICU stay, while the prediction window extends from the time of ICU admission to discharge. This approach leads to an observation window of 24 hours and a prediction window with varying duration, representing the length of stay in the ICU. Because MIMIC-III is composed of measurements taken at irregular time intervals, the input variables are aggregated hourly, resulting in a maximum of 24 input timestamps.

\end{document}